# Predicting population neural activity in the Algonauts challenge using end-to-end trained Siamese networks and group convolutions

Georgin Jacob, Harish Katti
Centre for Neuroscience, Indian Institute of Science, Bangalore-560012
georgin@iisc.ac.in, harishk@iisc.ac.in

## Introduction

The Algonauts challenge (Cichy et al., 2019) poses the problem of building models that can predict the brain data in the form of representational dissimilarity matrices (RDMs). The organisers provide fMRI as well as MEG RDMs for two training image sets (92, 118 images). Training RDMs are provided for early visual cortex (EVC) and inferior temporal cortex (IT) in fMRI, the organisers also provide MEG based RDMs for early and late responses. The training and test images include a diverse set of categories including faces, bodies and body-parts, tools, buildings, landscapes. Performance of the model is evaluated by predicting the RDMs of a test set of 78 images.

## Motivation

Deep convolutional networks trained using supervised methods have largely unseated all previously known algorithms in problems involving visual categorisation (Russakovsky et al., 2015). Our initial explorations with the Algonauts dataset revealed that there was some structure in the fMRI and MEG data. Yet, seemingly good choices of regression-based models that attempted to predict RDM distances using optimal combinations of distance matrices derived from pre-trained CNN layers, could explain only a small fraction of variance in the data. The pre trained networks that we used (AlexNet, VGG-face, VGG) had been trained to assign a category label to each input image. We then trained a series of Siamese models that take pairs of images as inputs and directly predict the RDM distance between a pair of images. We also iteratively improved model predictions by making novel choices for the model architecture as well as pre-processing the RDM training data.

## Methods

**Train and test data:** The training data comprised of two image sets with 92 and 118 images respectively. These were accompanied with fMRI and MEG activity RDMs corresponding to EVC and IT (fMRI) and early and late responses (MEG). After analysing the RDM matrices and some CNN training iterations, we found that CNNs trained with group averaged RDMs produced the best results.

**CNN architecture and training choices:** Siamese networks (Koch, Zemel, & Salakhutdinov, 2015) get their name from sharing of weights between two identically defined convolutional networks. Our network has a Siamese body and a custom head with group convolutions (Zhang, Qi, Xiao, & Wang, 2017). Body of our network comprises of convolutional layers from AlexNet (Krizhevsky, Alex, Ilya Sutskever, 2012) and the head consists of interleaved group convolutions (512 input x 32 outputs, 3x3 kernel, stride 1 and 16 groups), ReLU, Average Pool and linear layer (128 input x 1 output). Our network is shown in the following Figure 1.

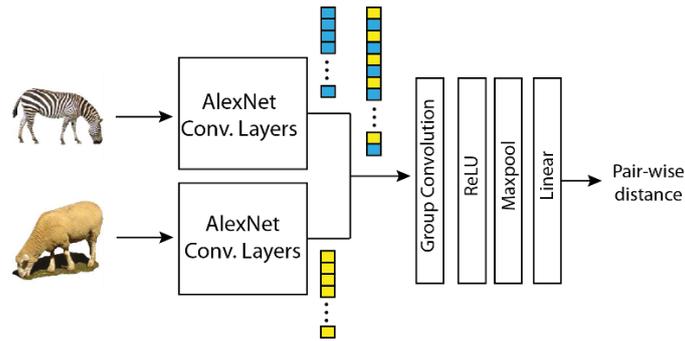

Figure 1: Schematic of our Siamese model architecture.

The weights of the network body were initialized with weights of a pretrained AlexNet and weights of the head was initialized randomly. The final distance prediction was a linear operation and we minimized the Euclidean loss with RDMs. Although a distance operation can be directly performed at Siamese stage, we found that it is more powerful to add an additional set of layers beginning with a group convolution that operates simultaneously on the two equal-length feature vectors derived from the input image pair. The entire network was implemented in python using the fast.ai package (www.fast.ai) in the jupyter notebook environment. Models were trained either on Nvidia GeForce GTX 1080 card running the Ubuntu18.04 operating system.

We first estimated an optimal learning rate using the cyclical learning rate algorithm (Smith, 2015) provided in fast.ai and then initiated 10-20epochs of training while keeping the body of the network frozen. Subsequently we ran close to 1000 training epochs after unfreezing the shared weights in the body of the network.

**RDM normalisation:** Since we were computing a Euclidean loss function, we normalized the training RDM distance values to match the range of our outputs. We found that this significantly improved model performance of our end-to-end trained CNNs.

**Results:** The performance of our Siamese CNNs is tabulated below for both the initial test set as well as the hidden test set that was released later. We report both the model correlation as well as explainable variance explained as provided to us by the organisers. Performance of the Siamese CNN in terms of explained variance (noise normalised correlation squared) is,

|  | fMRI | | MEG | |
| --- | --- | --- | --- | --- |
|  | EVC | IT | Early | Late |
| 78 images | 9.3% | 10.1% | 36.4% | 45.4% |
| Hidden test set |  |  | 18.1% | 27.3% |

Table 1: Results of our best performing model on the challenge and hidden test set.

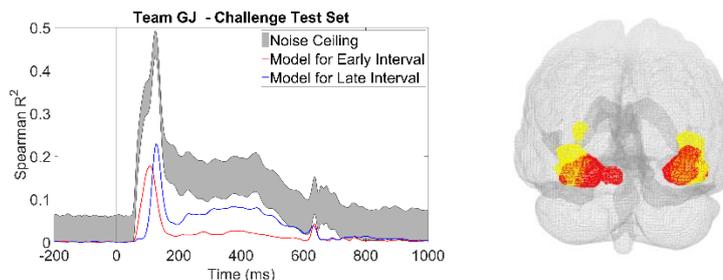

Figure 2: Left panel shows our predicted RDMs for early and late MEG activity translate correctly into an early and late peak when correlated with MEG sensor data (panel provided by organisers). The panel

on the right visualises that our predicted EVC (red) and IT (yellow) RDMs map to the correct part of the whole brain activity.

**Discussion:** Overall, we found that appropriately chosen deep convolutional networks can give the best account of the population level neural activity. We found that directly using pre-trained CNNs to predict neural distances was not as effective as end-to-end training to predict RDM distances. Pre-processing the population neural distances also improved the effectiveness of predictions.

**References**

Cichy, R. M., Roig, G., Andonian, A., Dwivedi, K., Lahner, B., Lascelles, A., … Oliva, A. (2019). *The Algonauts Project: A Platform for Communication between the Sciences of Biological and Artificial Intelligence*. *2*. Retrieved from http://arxiv.org/abs/1905.05675

Koch, G., Zemel, R., & Salakhutdinov, R. (2015). Siamese Neural Networks for One-shot Image Recognition. *ICML Deep Learning Workshop*, *2*. Retrieved from http://www.cs.cmu.edu/~rsalakhu/papers/oneshot1.pdf

Krizhevsky, Alex, Ilya Sutskever, and G. E. H. (2012). ImageNet Classification with Deep Convolutional Neural Networks. *Advances in Neural Information Processing Systems 25 (NIPS 2012)*, 1–9. Retrieved from https://papers.nips.cc/paper/4824-imagenet-classification-with-deep-convolutional-neural-networks

Russakovsky, O., Deng, J., Su, H., Krause, J., Satheesh, S., Ma, S., … Fei-Fei, L. (2015). ImageNet Large Scale Visual Recognition Challenge. *International Journal of Computer Vision*, *115*(3), 211–252. https://doi.org/10.1007/s11263-015-0816-y

Smith, L. N. (2015). *Cyclical Learning Rates for Training Neural Networks*. (April).

Zhang, T., Qi, G.-J., Xiao, B., & Wang, J. (2017). *Interleaved Group Convolutions for Deep Neural Networks*. (ii). Retrieved from http://arxiv.org/abs/1707.02725